\documentclass[runningheads]{llncs}
\usepackage{times}

\usepackage{soul}
\usepackage{url}
\usepackage[hidelinks]{hyperref}
\usepackage[utf8]{inputenc}
\usepackage[small]{caption}
\usepackage{graphicx}
\usepackage{amsmath}
\usepackage{amssymb}
\usepackage{booktabs}
\usepackage{bm}
\usepackage{fancyhdr}
\begin{document}

\title{Embed Wisely: An Ensemble Approach to Predict ICD Coding}
%
%
\author{Pavithra Rajendran\inst{1} \and
Alexandros Zenonos\inst{1} \and
Joshua Spear\inst{2} \and
Rebecca Pope\inst{3}
}
\authorrunning{P. Rajendran  et al.}
\institute{KPMG UK, 15 Canada Square, London E14 5GL, UK\\ \email{\{Pavithra.Rajendran,Alexandros.Zenonos\}@kpmg.co.uk} \and
Great Ormond Street Hospital for Children NHS Foundation Trust, Great Ormond Street, London WC1N 3JH, UK\\
\email{joshua.spear@gosh.nhs.uk} \and
UCL Great Ormond Street Institute of Child Health, 30 Guilford St, London WC1N 1EH, UK\\ \email{r.pope@ucl.ac.uk}
}
\maketitle              
\footnotetext[1]{The final authenticated version is available online at \href{https://doi.org/10.1007/978-3-030-93733-1_26}{https://doi.org/10.1007/978-3-030-93733-1\_26}.}%
\begin{abstract}
International Classification of Diseases (ICD) are the de facto codes used globally for clinical coding. These codes enable healthcare providers to claim reimbursement and facilitate efficient storage and retrieval of diagnostic information. The problem of automatically assigning ICD codes has been previously approached as a multilabel classification problem, using neural models and unstructured data. We utilise an approach for efficiently combining multiple sets of pretrained word embeddings to enhance the performance on ICD code prediction. Using post-processing and meta-embeddings techniques, we exploit the geometric properties of word embeddings and combine different sets of word embeddings into a common dimensional space. We empirically show that infusing information from biomedical articles, whilst preserving the local neighbourhood of the embedding, improves the current state-of-the-art deep learning architectures. Furthermore, we demonstrate the efficacy of this approach for a multimodal setting, using unstructured and structured information.
\end{abstract}

\section{Introduction}
\label{sec:intro}
The International Classification of Diseases (ICD) was created in 1893, when a French doctor named Jacques Bertillon named 179 categories of causes of death. It has been revised every ten years since then and has become an important standard for information exchange in the health care sector. Importantly, it has been endorsed by the World Health Organisation (WHO) and has been widely adopted by physicians and other health care providers for reimbursement, storage and retrieval of diagnostic information. 

ICD coding is the process of assigning ICD codes to a patient's condition. This, however, is an extremely complex process for several reasons: 
\begin{itemize}
    \item The label space of codes is large with over 15,000 codes in the ICD-9 taxonomy and 140,000 in the ICD-10-CM/PCS taxonomy. Furthermore, the codes are organised in a hierarchical structure where the top-level codes represent generic disease categories and the bottom-level codes represent more specific diseases. 
    \item In order to determine which codes to apply to a patient, a large amount of patient data, spread across different sources needs to be navigated and analysed. 
      
\end{itemize}


While ICD codes are important for making clinical and financial decisions, for the reasons mentioned above, clinical coding is time-consuming, error-prone and expensive. This clearly motivates the requirement for accurate, automated clinical coding which is explored in this paper.





Prior works exploring automated clinical coding using the MIMIC-III dataset have made use of deep learning methods. Specifically, assigning ICD codes to patients based on discharge summaries ~\cite{li2020icd,DBLP:conf/naacl/MullenbachWDSE18,xu2019multimodal}. However, dealing with the clinical text, for example, the text found in discharge summarise is challenging from an NLP perspective, as it includes irrelevant information, 
has an informal tone, does not necessarily follow correct grammatical conventions,  contains a large medical vocabulary and contains texts of highly varied lengths. Furthermore, diagnosis and textual descriptions of ICD codes written by clinicians can be written in differing styles despite referring to the same disease. As a result, ICD code definitions have also been used to enrich the label information which, unlike the summaries, are formally and precisely worded.
\newline

In our work, we propose a novel multimodal approach, utilising information present in unstructured data (i.e. discharge summaries) and structured data (such as heart rate, haemoglobin, respiratory rate) and develop a meta-classifier for ensembling the predictions of various models, trained on different modalities of data. Furthermore, we utilise a novel approach for combining multiple sets of pretrained word embeddings to improve performance on unstructured data. We empirically demonstrate that the final ensemble model outperforms the baseline method in the multi-label classification task of ICD 10 and ICD 9 coding.
\newline

Pretrained word embeddings have been successfully used in downstream NLP tasks ~\cite{DBLP:conf/emnlp/Kim14} since they capture semantic relationships and provide better text representations. In order to further improve these representations, we consider incorporating external knowledge from separate datasets without increasing the dimensionality of the data used for training (augmenting the dataset). Instead
we focused on exploiting the geometric properties of the pretrained word embeddings as well as, combining embeddings trained on external knowledge into a common dimensional space using using meta-embedding techniques~\cite{DBLP:conf/ijcai/BollegalaHK18,DBLP:conf/naacl/CoatesB18,DBLP:conf/emnlp/KielaWC18}. We hypothesise that this approach would be particularly useful for instances where access to pretrained word embeddings is available but the ability it derive them is not (i.e. through lack of significant compute or lack of access to the underlying raw data). To study the potential advantage of our proposed approach, we consider prior work~\cite{li2020icd,DBLP:conf/naacl/MullenbachWDSE18} utilising pretrained word embeddings trained on the MIMIC dataset. 
\newline

The contributions of this paper are as follows:
\begin{description}
    \item [1] Our first contribution is a simple and effective approach for infusing external knowledge into word embeddings derived from textual summaries, without augmenting the dimensionality of the dataset used for training. The semantic information captured by the textual summaries is not necessarily the same as the semantic information captured by the external knowledge. Therefore, in order to preserve the rich representation from both sources, we focus on techniques for infusing the knowledge of both sources without drastically increasing the complexity of the model. This is achieved by exploiting the geometric properties of word vectors and combining embeddings trained on external information using meta-embedding techniques. We empirically evaluate different approaches using the top 32 ICD 10 and 50 ICD 9 codes and our results indicate that the best performance is obtained by preserving the local neighbourhood of the word vectors while combining them into meta-embeddings.
    
    \item [2] Our second contribution is an ensemble approach that utilises various modalities, combining both structured and unstructured information. We utilise information from various sources to enrich our dataset with information that is potentially missing from the summaries. Our ensemble model empirically outperforms the baseline methods.
\end{description}

\section{Related Work}
\label{sec:related}

There has been a growing interest within the field of natural language processing for learning representations of words as vectors, also known as word embeddings. These embeddings have been derived via two core methods, namely count-based and \\ prediction-based methods. The Glove algorithm~\cite{pennington2014glove} is a popular count-based approach that makes use of the co-occurrences probabilities of words. The Word2Vec~\cite{mikolov2013distributed} approach is a popular prediction-based approach, in which models are trained based on CBOW (continuous bag of words) or the Skip-Gram approach. A growing interest in topics related to word embeddings is reflected by the numerous
papers published in various NLP conferences. 

Due to vast amounts of data and compute required to train embedding models, many practitioners have utilised pretrained word embeddings. Pretrained word embeddings (or static word embeddings) refer to word embeddings which do not change in a given context. A new area of interest within word embeddings is to learn \emph{meta-embeddings} from multiple sets of pretrained word vectors without having the underlying text sources on which the embeddings have been trained. One of the simplest approaches proposed has been to use concatenation followed by averaging~\cite{DBLP:conf/naacl/CoatesB18}. Yin et al.~\cite{yin2016learning} was one of the earliest works to investigate meta-embeddings. They used a projection layer known as 1TON for computing the meta-embeddings in a linear transformation.
However, few works have studied the use of meta-embeddings for the healthcare domain~\cite{DBLP:conf/biostec/ChowdhuryZYL20,el2019embedding}. To the best of our knowledge, the application of meta-embedding techniques for multilabel classification of ICD coding has not yet been investigated. In this work we investigate several different approaches for doing so, in particular, we focus on combining pretrained in-domain word embeddings with pretrained word embeddings derived from external sources such as scientific articles on the Web.


Automatic ICD coding using unstructured text data has been explored by researchers for several years whereby the full breadth of learning approaches have been considered. Koopman et al.~\cite{koopman2015automatic} utilised a multi-label classification approach and combined SVM classifiers via a hierarchical model to assign ICD codes to patient death certificates, first identifying whether the cause of death was due to cancer, then identifying the type of cancer and associated ICD code. Whilst the final solution performed well, there were two key limitations to the approach. Firstly, the coverage of cancers included in the dataset was very imbalanced resulting in cancer types associated with rarer diseases being harder to predict. Secondly, the cancer identification model was susceptible to false-positives when a patient was cited as having cancer but it was not the primary cause of death. 

More recently, the scope of the problem has been extended to include multiple ICD codes and it has been addressed via multi-label methods with researchers more often utilising deep learning approaches, generally centred around CNN and LSTM based architectures. Mullenbach et al.~\cite{DBLP:conf/naacl/MullenbachWDSE18} adopted a CNN architecture with a single filter, defining a per-label attention mechanism to identify the relevant parts of the latent space. The CNN architecture~\cite{DBLP:conf/emnlp/Kim14} has been proven to be useful for sentence classification tasks and in this work, they empirically show that CNN is better for ICD code prediction. The per-label attention provided a means of scanning through the entire document without limiting it to a particular segment of the data. This approach achieved state-of-the-results across several MIMIC datasets. In our approach, we use a similar CNN-based architecture but instead focus on infusing external information via embedding vectors and combining structured features, along with the unstructured information. Both approaches are explained in the paper along with empirical evidence of the improved performance that our proposed approach gives.

Vu et al.~\cite{ijcai2020-461} proposed a BiLSTM encoder along with a per-label attention mechanism, inspired by the work of Lin et al.~\cite{DBLP:conf/iclr/LinFSYXZB17} and proved it to perform well at generating general sentence embeddings. Here, Vu et al.~\cite{ijcai2020-461} extended the attention mechanism by generalising it for multilabel classification by performing an attention hop per label.



Xie et al.~\cite{xu2019multimodal} proposed a text-CNN for modelling unstructured clinician notes however, rather than implement an attention mechanism, the authors extracted features via TF-IDF from the unstructured guidelines provided to professionals when defining the ICD classifications. By including these features along with the convolved CNN layer, the authors mimicked the input that the professionals would get from the ICD coding guidelines. To enrich the predictions of the unstructured data model the authors used an ensemble-based method of three models. Semi-structured data was utilised by embedding the ICD code descriptions in the same latent vector space as the diagnosis descriptions and structured data was utilised through a decision tree model. The imbalance issues in the data were addressed via Label Smoothing Regularization and the resulting model achieved state-of-the-art accuracy, for the time, as well as improving model interpretability. However, they did not disclose what structured data they used specifically and what features were used.


Shi et al.~\cite{Shi2017TowardsAI} proposed a hierarchical deep learning model with attention mechanism that automatically assigned ICD diagnostic codes given a written diagnosis. They also proposed an attention mechanism that was designed to address the mismatch between diagnosis description number and assigned code number. The results showed that the soft attention mechanism improved performance. However, they only focused on the top 50 ICD-9 codes. 

\section{Proposed Approach}
\label{sec:proposed}
\begin{figure}
    \centering
    \includegraphics[width=10cm]{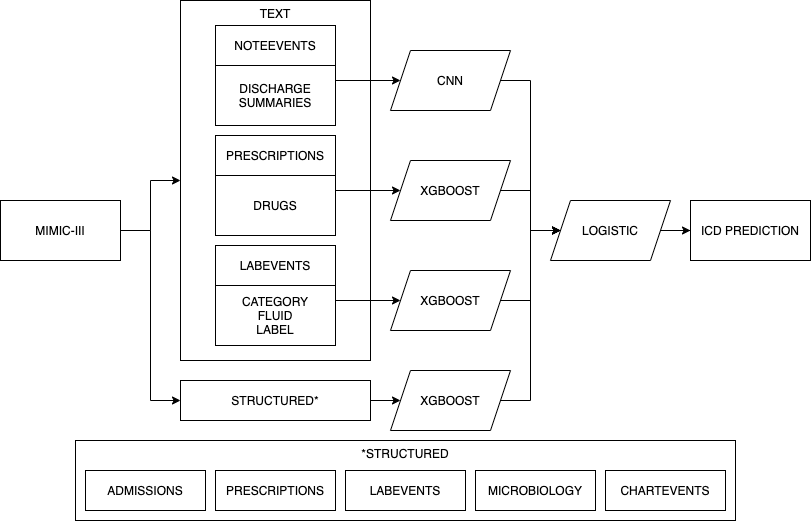}
    \caption{Multimodal ensemble architecture for ICD prediction}
    \label{fig:architecture}
\end{figure}

In this section, we explain our proposed multi-label classification approach with discharge summaries (unstructured data) as well the multimodal approach (structured and unstructured data) for automatically predicting the ICD coding (Figure.~\ref{fig:architecture}). 

\subsection{Unstructured Data}
\label{sec:proposed_unstructured}
In this subsection, we explain our approaches for classifying unstructured data i.e., discharge summaries. Based on prior work~\cite{DBLP:conf/naacl/MullenbachWDSE18}, we follow a similar deep learning approach using a convolutional neural network with a per-label attention mechanism that can span across the entire document to identify the portions of the document that correspond to the different labels. 

Differing from~\cite{DBLP:conf/naacl/MullenbachWDSE18}, we propose to study the effect of post-processing word vectors to understand whether we can capture the semantic representations of word embeddings effectively. Furthermore, we investigate the impact of utilising external data from PubMed articles to aid with the ICD classification. However, we propose to infuse this external information into discharge summaries via a meta-embedding technique rather than augmenting the training data with the PubMed articles.


For the rest of the section, we assume that we have access to trained word embeddings from MIMIC discharge summaries using the Word2Vec~\cite{mikolov2013distributed} algorithm as (1) our purpose here is to focus on exploiting as much of the semantic information captured as possible, for better utilisation of embeddings and (2) it serves as a baseline comparison with prior work~\cite{DBLP:conf/naacl/MullenbachWDSE18}. The details of the different post-processing steps~\cite{DBLP:conf/iclr/MuV18} are described below.

\subsubsection{Post-processing Word Vectors}
\label{sec:postprocess}
Let us consider a set of words present in a corpus, represented as  $\textrm{w} \in \textrm{V}$, such that each word $\textrm{w}$ is represented by a pretrained word embedding $\boldsymbol{w}_{i} \in \mathbb{R}^{k}$ in some $k$ dimensional vector space. We term the first post-processing step ~\cite{DBLP:conf/iclr/MuV18} as
$\textbf{MeanDiff}$. The $\textbf{MeanDiff}$ step is implemented by first computing the mean embedding vector, $\hat{\boldsymbol{w}}$, of all words in $\textrm{V}$ and subsequently subtracting $\hat{\boldsymbol{w}}$ from all words in $\textrm{V}$. The process is defined as follows:

 \begin{align}
  \hat{\boldsymbol{w}} = \frac{1}{|\textrm{V}|} \sum_{\textrm{w} \in \textrm{V}} \boldsymbol{w}_{\textrm{w}}; \ \ \  \tilde{\boldsymbol{w}}_{\textrm{w}} =  \boldsymbol{w}_{\textrm{w}} - \hat{\boldsymbol{w}};
  \forall_{\textrm{w} \in \textrm{V}}
 \label{eq:mean_subtracted}
 \end{align}
 
Mu and Viswanath~\cite{DBLP:conf/iclr/MuV18} observed that the normalised variance ratio of principal components of word embeddings decayed until some top $l \leq d$ components and remained constant after that. They proposed first applying the $\textbf{MeanDiff}$ to a set of embedding vectors $\boldsymbol{w}_{\textrm{w}}$ such that $\textrm{v} \in \textrm{V}$ to obtain a processed set of embedding vectors. They then removed the top $l$ principle components from the embedding vectors, with the hypothesis that the majority of the word specific semantic information is captured in the remaining $d-l$ components. Based on the above observation, the second post-processing step is to apply principal component analysis to the word vectors obtained using Eq.~\ref{eq:mean_subtracted} and to remove the first $l$ principal components from each individual word vector. We term this step as $\textbf{PCADiff}$ for our usage. To perform $\textbf{PCADiff}$ we arrange the word vectors as columns within a matrix $\textbf{A} \in \mathbb{R}^{k \times |\textrm{V}|}$ and subsequently find the principal component $\boldsymbol{u}_{1}, \ldots, \boldsymbol{u}_{d}$. For each word vector $\boldsymbol{w}_{\textrm{w}}$, we then remove the first $l$ principal components as follows:

 \begin{align}
 \label{wordpca}
 \boldsymbol{w}'_{\textrm{w}} = \tilde{\boldsymbol{w}}_{\textrm{w}} - \sum_{i=1}^{l} \left( \boldsymbol{u}^{\top}_{i} \boldsymbol{w}_{\textrm{w}} \right) \boldsymbol{u}_{i} 
 \end{align}

In the above approach, we focus on in-domain trained word embeddings i.e the word vectors trained on the MIMIC discharge summaries.

The next section discusses the different meta-embedding techniques for combining word vectors trained on external knowledge with the in-domain trained word embeddings. Recent work~\cite{yin2016learning} demonstrated that pretrained word vectors trained on the same source of information but with different algorithms varied in the semantics captured. We propose that the same intuition can be applied to using the same algorithm but on different datasets. In order to capture the semantics from both sets of word embeddings i.e MIMIC discharge summaries and the PubMed articles, we combine the information in the form of "\emph{meta-embeddings}" i.e combining the different embeddings into a common meta-embedding space. While there are several approaches to achieve meta-embeddings, in this paper we focus on two methodologies which are described below.

\subsubsection{Meta-embeddings}
\label{sec:meta_embeddings}
Let us assume, for simplicity, that we have two different sources of information i.e one based on the discharge summaries and the other based on the PubMed articles. Let us represent the vocabulary of words present in the two sources of information as $\mathcal{S}_{1}$ and $\mathcal{S}_{2}$, respectively. 

The two methodologies for obtaining meta-embeddings are explained below.

\paragraph{Averaging Meta-embedding} Here, we assume that the words present in $\mathcal{S}_{1}$ and $\mathcal{S}_{2}$ are trained with the same algorithm i.e Word2Vec to produce word vectors.  Furthermore, we assume that the vectors belonging to both sources have a common dimensionality, $\textrm{k}$. Prior work~\cite{DBLP:conf/naacl/CoatesB18} showed that averaging the embedding vectors was a useful meta-embedding technique that achieved comparable results to that of concatenating the vectors.

To understand this, let us consider the meta-embedding of each word, represented by averaging the word vectors from the two sources as follows:
\begin{equation}
    \hat{\boldsymbol{w}} = \frac{\boldsymbol{w}_{\mathcal{S}_{1}} + \boldsymbol{w}_{\mathcal{S}_{2}}}{2}
    \label{eq:avg_meta}
\end{equation}

In this case, the Euclidean distance between two words $\hat{\boldsymbol{w}}_{1}$ and $\hat{\boldsymbol{w}}_{2}$ based on Eq.~\ref{eq:avg_meta} is given as follows:
\begin{equation}
    \textrm{E}_\textrm{AVG} =\left\Vert \hat{\boldsymbol{w}}_{1} - \hat{\boldsymbol{w}}_{2} \right\Vert_2  = \frac{1}{2} \left\Vert \frac{\boldsymbol{w}_{1S_1} + \boldsymbol{w}_{1S_2}}{2} - \frac{\boldsymbol{w}_{2S_1} + \boldsymbol{w}_{2S_2}}{2} \right\Vert_2
\label{eq:meta_avg_final}
\end{equation}
Coates and Bollegala~\cite{DBLP:conf/naacl/CoatesB18} showed that the source embeddings were approximately orthogonal  and hence, averaging captured approximately similar information as concatenation, without increasing the dimensionality. 
Hence, we use this averaging method for combining the word embeddings as meta-embeddings.
\newline

\paragraph{Locally Linear Meta-embedding} In the above method, we find that averaging-based meta-embeddings performs comparably to
concatenation ~\cite{DBLP:conf/naacl/CoatesB18}. However, one of it's limitations is that it does not capture the variations present within the local neighbourhood of the word vectors in their respective sources. To address this issue, Bollegala et al.~\cite{DBLP:conf/ijcai/BollegalaHK18} constructed meta-embeddings in an unsupervised manner where they considered the mapping of embeddings from different sources to a common meta-embedding space based on the local neighbourhood of a word in each of the sources.
Two steps were performed (1) the word embeddings in each source were reconstructed based on their local neighbourhoods (2) these reconstructions were used to project the embeddings into a common meta-embedding space such that the the local neighbourhoods were preserved.
\newline

Let us assume we have a set of words present in the two sources i.e. $\mathcal{S}_{1}$ and $\mathcal{S}_{2}$. Let us also represent their vocabularies as $\mathcal{V}_{1}$ and $\mathcal{V}_{2}$ respectively such that, $\mathcal{V}$ represents the common set of vocabulary i.e $\mathcal{V}_{1} \cap \mathcal{V}_{2}$. Further, for each word $\textrm{v} \in \mathcal{V}_{1}$, we represent the word vector as $\boldsymbol{v}_{\mathcal{V}_{1}} \in \mathbb{R}^{\textrm{d}_{1}}$. Similarly, for each word $\textrm{v} \in \mathcal{V}_{2}$, we represent the word vector as $\boldsymbol{v}_{\mathcal{V}_{2}} \in \mathbb{R}^{\textrm{d}_{2}}$. Here, $\textrm{d}_{1}$ and $\textrm{d}_{2}$ represent the dimensions of the vectors respectively.

In the reconstruction step, for each word $\textrm{v} \in \mathcal{V}$ i.e $\textrm{v} \in \mathcal{V}_{1} \cap \mathcal{V}_{2}$, we obtain the \emph{k} nearest neighbours present within the two sources $\mathcal{S}_{1}$ and $\mathcal{S}_{2}$. This is carried out using the BallTree algorithm based on prior work~\cite{DBLP:conf/ijcai/BollegalaHK18} since this approximate methodology reduces the time complexity in identifying the approximate \emph{k} neighbours. Let us denote the neighbours as $\mathcal{N}_{\mathcal{V}_{1}}$ and $\mathcal{N}_{\mathcal{V}_{2}}$ respectively. Furthermore, we denote the neighbours of word $v$ in $\mathcal{N}_{\mathcal{V}_{i}}$ as $\mathcal{N}_{\mathcal{V}_{i}}(v)$.

\begin{equation}
\Psi(\text{\bf{W}}) = \sum_{i=1}^{2} \sum_{\textrm{v} \in \mathcal{V}} \left\Vert \boldsymbol{v}_{\mathcal{V}_{i}} - \sum_{ \textrm{u} \in \mathcal{N}_{\mathcal{V}_{i}} (\textrm{v})} \textrm{w}_{\textrm{u}\textrm{v}}\boldsymbol{u}_{\mathcal{V}_{i}} \right\Vert^{2}_{2} 
\label{eq:reconstuction}
\end{equation}

where
$\textrm{w}_{\textrm{u}\textrm{v}} = 0;$ if the words are not \emph{k} neighbours in either of the sources.

\begin{equation}
\begin{split}
& \frac{\partial(\Psi(\text{\bf{W}}))}{\partial\textrm{w}_{\textrm{u}\textrm{v}}} = \\
& -2 \sum_{i=1}^{2}\left( \boldsymbol{v}_{\mathcal{V}_{i}} -  \sum_{\textrm{x}\in \mathcal{N}_{\mathcal{V}_{\textrm{i}}}(\textrm{x})} \textrm{w}_{\textrm{v}\textrm{x}}\boldsymbol{x}_{\mathcal{V}_{i}} \right)
^\top \boldsymbol{u}_{\mathcal{V}_{i}}\mathbb{I} [\textrm{u} \in \mathcal{N}_{\mathcal{V}_{i}}(\textrm{v})]
\end{split}
\label{eq:meta_error_grad}
\end{equation}

Given a word $\textrm{v} \in \mathcal{V}$, for each neighbouring word $\textrm{u}$ in $\mathcal{N}_{\mathcal{V}_{1}}$ and $\mathcal{N}_{\mathcal{V}_{2}}$, the recontruction weights are learned such that the reconstruction error given in Eq.~\ref{eq:reconstuction} is minimized i.e minimizing the sum of the local distortions present in the two sources. To achieve this, the error gradient is computed using Eq.~\ref{eq:meta_error_grad}.

Based on prior work~\cite{DBLP:conf/ijcai/BollegalaHK18}, the weights are uniformly randomly initialized for each \emph{k} neighbour and optimal weights are obtained using stochastic gradient descent (SGD) with the initial learning rate set as 0.01 and maximum number of iterations set to 
100.

The weights are normalized and used in the projection step.
The projection step makes use of the normalized reconstructed weights and learns the meta-embeddings of the words $\textrm{u}, \textrm{v} \in \mathcal{V}$ in a common dimensional space $\mathcal{P}$ i.e $\boldsymbol{u}_{\mathcal{P}}, \boldsymbol{v}_{\mathcal{P}} \in \mathbb{R}^{d_{\mathcal{P}}}$ such that 
the local neighbourhood from both the sources is preserved.  This is done using a truncated eigensolver, minimising the projection cost given below:

\begin{equation}
\Psi(\mathcal{P}) = \sum_{i=1}^{2} \sum_{\textrm{v} \in \mathcal{V}} \left\Vert \boldsymbol{v}_{\mathcal{P}} - \sum_{ \textrm{u} \in \mathcal{N}_{\mathcal{V}_{i}} (\textrm{v})} \textrm{w}^{\prime}_{\textrm{u}\textrm{v}}\textbf{u}_{\mathcal{P}} \right\Vert^{2}_{2} 
\label{eq:meta_lle}
\end{equation}
where
\begin{equation}
\textrm{w}^{\prime}_{\textrm{u}\textrm{v}} = \textrm{w}_{\textrm{u}\textrm{v}} \sum_{i=1}^{2} \mathbb{I}[\textrm{u} \in \mathcal{N}_{\mathcal{V}_{i}} (\textrm{v})]
\label{eq:lle_expand}
\end{equation}
such that if $\textrm{x}$ = 1; $\mathbb{I}(\textrm(x))$ = \emph{True} and \emph{False} otherwise.

The meta-embeddings are obtained by computing the smallest $(\textrm{d}_{\mathcal{P}} + 1)$ eigenvectors of the matrix given below.

\begin{equation}
 \text{\bf{M}} = (\text{\bf{I}} - \text{\bf{W}}^{\prime})^\top(\text{\bf{I}} - \text{\bf{W}}^{\prime})
 \label{eq:lle_eigen}
\end{equation}
such that matrix $\text{\bf{W}}^{\prime}$ contains the values computed using Eq.~\ref{eq:meta_lle}.

Different variants of pretrained word embeddings are obtained based on the above discussed methodologies. These different embedding vectors are then fed into a neural model for multilabel classification of ICD codes.

For the purpose of comparison, we use the CNN-based architecture following prior work~\cite{DBLP:conf/naacl/MullenbachWDSE18} which is described below.

\subsubsection{CNN Encoder} A document is represented as $\textrm{X} = \{ \boldsymbol{x}_{1}, \ldots, \boldsymbol{x}_{N}\}$ such that each word is represented using a pretrained word vector. A convolutional neural architecture is used to encode the document at each n step as:

\begin{equation}
    \boldsymbol{h}_{i} = \tanh(\text{\bf{W}}_{c}\ast \boldsymbol{x}_{\textrm{i}:\textrm{i}+\textrm{k}-1} + \textrm{b}_\textrm{c})
    \label{eq:cnn}
\end{equation}

where $\text{\bf{W}}_{c} \in \mathbb{R}^{\textrm{k} \times \textrm{d}_{\textrm{e}} \times \textrm{d}_{\textrm{c}}}$ represents the convolutional filter such that $\textrm{k}, \textrm{d}_{\textrm{e}}, \textrm{d}_{\textrm{c}}$ denote filter width, input embedding dimension and filter output size respectively.

\subsubsection{Per-label Attention} Following the base representation of the document as \\ $\text{\bf{H}} = \{\textrm{h}_{1}, \ldots, \textrm{h}_{N}\}$, for each label $\textit{l}$, a per-label attention vector is computed as follows:
\begin{equation}
\boldsymbol{a}_{\textit{l}} = Softmax(\text{\bf{H}}^T\boldsymbol{u}_{\textit{l}})
\label{eq:per_label_attention}
\end{equation}
where $\boldsymbol{u}_{\textit{l}}$ represents the vector parameter for each label $\textit{l}$.
This attention vector is then used for defining the label-based document representations as follows:
\begin{equation}
    \boldsymbol{v}_{\textit{l}} = \sum_{n=1}^{\textrm{N}} \boldsymbol{a}_{\textit{l},\textit{n}}\boldsymbol{h}_{\textit{n}}
    \label{eq:attn_vec}
\end{equation}

\subsubsection{Classification} The probability per label is computed as follows:
\begin{equation}
    \hat{\textrm{y}}_{\textit{l}} = \sigma(\alpha_{\textit{l}}^{\top}\boldsymbol{v}_{\textit{l}} + \textrm{b}_{\textit{l}})
    \label{eq:cnn_classification}
\end{equation}
where $\alpha \in \mathbb{R}^{d_{c}}$ represents the vector containing the prediction weights. 

The training objective aims to minimize the binary-cross entropy loss as given below.
\begin{equation}
\textrm{Loss}(\textrm{X}, \textrm{y}) = - \sum_{\textit{l} = 1}^{\mathcal{L}} \textrm{y}_{\textit{l}}\log(\hat{\textrm{y}}_{\textit{l}}) + (1 - \textrm{y}_{\textit{l}})\log(1 - \hat{\textrm{y}}_{\textit{l}})
\label{eq:cnn_loss}
\end{equation}

The probability scores per label are further used in combination with structured data to improve the performance of the multilabel classification, which is explained in detail in the following subsection.

\subsection{Multimodal Approach}
\label{sec:multimodal}
We utilise structured data from the MIMIC-III dataset that is spread across several tables and contains information about a patient's care during their hospital admission as well as some textual information like medication and lab results. The specific information we use relates to admission information, lab reports, prescriptions, vital signs (chart events table) and microbiology test results. 

\subsubsection{Structured Data}
\paragraph{Numeric Data} We aggregate the data up to an admission level and extract statistical properties, i.e., mean, standard deviation, min and max for each numeric value as well as the number of measurements taken per admission. 
In particular, we utilise the 100 most common items measured across all patients. These include, \emph{heart rate}, \emph{hemoglobin}, \emph{respiratory rate}, \emph{creatine}, \emph{bun}, \emph{wbc}, \emph{magnesium} etc.

\paragraph{Categorical Data}
\label{sec:multimodal_categorical}
Textual information contained in tables but with no particular meaning, i.e., medication (drugs) and laboratory exams are represented using Term Frequency-Inverse Document Frequency (TF-IDF) based features. 


\section{Experiments}
\label{sec:experiments}
This section provides a brief overview of the MIMIC-III dataset that we used in our experiments and the different experimental settings for both the unstructured data i.e discharge summaries and the multimodal data i.e. structured and unstructured information.
\subsection{Data and Preprocessing}
\label{sec:data}
\begin{table*}
\centering
  \caption{Total number of samples in train, test and dev set based on 32 ICD-10 codes and 50 ICD-9 codes respectively}
  \label{tab:samples}
  \begin{tabular}{cccl}
    \toprule
    Data & Train & Dev & Test\\
    \midrule
    32 ICD-10 & 28201 & 3134 &12430 \\
    50 ICD-9 & 8044 & 804 & 1725 \\
  \bottomrule
\end{tabular}
\label{datasplit}
\end{table*}

In this paper we used the well-known MIMIC-III database for empirically evaluating our approach. The dataset contained electronic health records of 58,976 patients who stayed in the Intensive Care Unit (ICU) of Beth Israel Deaconess Medical Centre from 2001 to 2012~\cite{mimiciii}. This included information such as demographics, vital sign measurements, laboratory test results, procedures, medications, caregiver notes and imaging reports. For our purpose, we removed admissions that were not associated with a discharge summary. Also, we removed patient admissions that did not have information on the admission itself (i.e. length of stay), laboratory results and prescriptions, as we required at least some monitoring of the patients health during their stay.
The total number of unique admissions after filtering was $44,765$.
In this work, our aim was to predict the ICD code classification for each of the admissions we were considering. Even though the MIMIC-III dataset included ICD-9 mappings, we manually mapped those to the new ICD-10 codes and focused on predicting only the top 32, similarly to~\cite{xu2019multimodal}. Also, for completeness we experimented with the top 50 ICD-9 codes as in previous studies.
A discharge summary was considered very useful for understanding what happened during an admission as it included but was not limited to information about the history of illness, past medical history, medication, allergies, family and social history, physical exam at the point of admission, lab result summary, procedures, discharge condition and status as well as discharge medication, follow-up plans, final diagnosis and other discharge instructions. In terms of preprocessing the summaries, we removed symbols and numbers not associated with a text. 
Table.~\ref{datasplit} contains the number of samples present within the training, development and test set for experimentation purposes.

\subsection{Experimental Settings}
\label{sec:experimental_settings}

Experiments based on both structured and unstructured data were carried out with the top 32 ICD-10 codes and top 50 ICD-codes in order to align with benchmarks in the literature.

\subsubsection{Unstructured Data}
\label{sec:experiment_unstructured}
Experiments based on unstructured information were carried out by extracting the discharge summaries based on the top 32-ICD codes as well as top 50 ICD-9 codes, separately. In addition, we extracted full scientific articles from PubMed, which totalled 672,589 articles ~\cite{moen2013distributional}.

The CNN-based architecture (also known as CAML)~\cite{DBLP:conf/naacl/MullenbachWDSE18} served as a baseline for comparison in experiments, since the architecture was used as a base model for all of the experiments we conducted. For the baseline experiment, we initialised the CAML architecture with embeddings trained on the MIMIC train dataset. The hyperparameters set were: embedding dimension as \emph{200}, dropout as \emph{0.5}, filter size as \emph{4}, learning rate as \emph{0.001}, batch size as \emph{16}, filter maps as \emph{50} and patience as \emph{3}. We trained the model for \emph{100} epochs with an early stopping criteria based on the \emph{micro-F1} score such that the training was stopped if the \emph{micro-F1} score did not improve after \emph{3} epochs.

We also experimented with the MultiResCNN architecture using the default hyperparameters~\cite{li2020icd} aside from embedding size, for which we used {200}. Due to the complexity of the model, we reported the scores based on the same epoch as the best epoch achieved using the baseline CNN-architecture (CAML). Based on this, we found the baseline model to perform better. As a result, we used the CAML architecture for the remaining experiments. In order to understand the adaptability of our proposed methodology, we also reported the scores of the most effective 
meta-embedding technique/MultiResCNN combination that we tried.



The baseline experiment reproduced the results based on prior work i.e the per-label attention based CNN architecture~\cite{DBLP:conf/naacl/MullenbachWDSE18}. To replicate these results, we used Word2Vec to generate word embeddings from the MIMIC training dataset and used them as an input to train the CNN neural network.

We investigated the different techniques explained in Section.~\ref{sec:proposed_unstructured} to obtain different input embeddings to train the neural model. Word embeddings trained on MIMIC discharge summaries and those trained on PubMed scientific articles are termed as $\textbf{W2V-MIMIC}$ and $\textbf{W2V-PubMed}$ respectively for our usage.

\begin{enumerate}
 \item \textbf{MeanDiff-Word2Vec-MIMIC} Mean vector of the vectors in $\textbf{W2V-MIMIC}$ removed from individual word vectors (Section.~\ref{sec:postprocess}).
\item\textbf{MeanDiff-PCADiff-Word2Vec-MIMIC} Above steps followed and extended by computing the principal components and removing the first $\textit{l}$ principal components from the mean removed word vectors. In our experiments, we considered $l$ = 2~\footnote{Prior work theoretically demonstrated that choosing l depends on the length of the embeddings} based on the embedding dimension i.e 200 which is chosen based on best performance.
\item \textbf{Averaging} Meta-embeddings obtained by combining ($\textbf{W2V-MIMIC}$) and \\ ($\textbf{W2V-PubMed}$) based on the technique explained in Section~\ref{sec:meta_embeddings}. Both the vectors had dimensions of 200.
\item \textbf{Locally Linear} Meta-embeddings obtained by combining ($\textbf{W2V-MIMIC}$) and ($\textbf{W2V-PubMed}$) based on the technique explained in Section.~\ref{sec:meta_embeddings}. The nearest neighbours for each source was set to 1200 since this gave the best result based on prior work~\cite{DBLP:conf/ijcai/BollegalaHK18}. Here, the dimensions were set to 200.
\item \textbf{MeanDiff-Averaging} First, steps in (1) were carried out on $\textbf{W2V-MIMIC}$ and $\textbf{W2V-PubMed}$ separately and then, combined using the technique explained in (3). 
\item \textbf{MeanDiff-PCADiff-Averaging} First, steps in (2) were carried out on \\ $\textbf{W2V-MIMIC}$ and $\textbf{W2V-PubMed}$ separately and then, combined using the technique (3).
\item \textbf{MeanDiff-Locally Linear}  First, steps in (1) were carried out on  $\textbf{W2V-MIMIC}$ and $\textbf{W2V-PubMed}$ separately and then, combined using the technique (4).
\item \textbf{MeanDiff-PCADiff-Locally Linear}
First, steps in (2) were carried out on \\ $\textbf{W2V-MIMIC}$ and $\textbf{W2V}$ $\textbf{-PubMed}$ separately and then, combined using the technique (4).
\end{enumerate}

Experiments were conducted using the same hyperparameters to train the baseline approach to benchmark our results against prior work.

    

\subsubsection{Multimodal Data}
\label{sec:experiment_multimodal}
For each experiment utilising structured data we trained 3 separate XGBoost models that utilised the different structured datasets available. For each experiment, we ran a randomised search for the best hyperparameters in a crossvalidation fashion. Their corresponding values are shown in Table~\ref{tab:xgboost32}.

\begin{enumerate}
\item \textbf{XGBoost Numeric Data}
This model utilised only numeric information available in the structured data.

\item \textbf{XGBoost Prescription Data}
This model utilised only information from the prescription table in the MIMIC-III dataset. In particular, we aggregated the drugs a patient received during an admission to a single row and applied TF-IDF to extract features.

\item \textbf{XGBoost Lab Exam Data}
This model utilised only lab exam data from the MIMIC-III dataset. Similar to the prescription data,
we aggregated the data to a single row and applied TF-IDF to extract features. 

\end{enumerate}

All three models were trained separately but on data concerning the same patients. All models made predictions on the same Test set as shown in Table~\ref{datasplit}.

\begin{table*}
\centering
  \caption{Hyperparameters used for the XGBoost Algorithm for the top 32 ICD-10 and top 50 ICD-9 experiments}
  \label{tab:xgboost32}
  \begin{tabular}{cll}
    \toprule
    Hyperparameter & 32 ICD-10 & 50 ICD-9) \\
   \midrule
    Colsample by tree & 0.85 & 0.98 \\
    Gamma & 0.86 & 0.78\\
    Subsample & 0.66 & 0.67\\
    Number of estimators & 2000 & 2000\\
    Max Depth & 5 & 7\\
    Min Child Weight & 5 & 4\\
    Learning rate & 0.15 & 0.19\\
  \bottomrule
\end{tabular}
\end{table*}


\subsubsection{Meta Data}
\label{sec:metadata}
As illustrated by Table~\ref{tab:samples}, we split the MIMIC dataset into three sets. The Train set was used for training the unstructured neural model, and the three structured XGBoost models in the multimodel experiments. The Dev set was used to perform early stopping during the training of the neural model. In the experiments that only concerned unstructured data, the Test set was used for reporting our final results, as shown in tables~\ref{resultunstructured32} and~\ref{resultunstructured50}. However, in the multimodel experiments, we performed 5-Fold cross validation on the Test set in order to train the meta classifier and report our final results. For each iteration of the k-fold procedure, predictions were made by the base models on 4 of the 5 folds which were subsequently used to train the meta-classifier. The trained meta-classifer then made predictions on the unseen 5th fold. The meta classifier's predictions on each of the 5 unseen holdout sets were then averaged and used to report the results in tables~\ref{resultstructured32} and~\ref{resultstructured50}. This approach was taken to ensure that the meta-classifier was trained using data that was unseen by the base models and therefore reduced the potential of overfitting. It also meant that the results reported for the multimodel experiments were still generated using unseen data.  



\section{Results}
\label{sec:results}
\begin{table*}

\caption{Micro and macro results are presented for multilabel classification of 32 ICD-10 CODES on the test set. \emph{Input Embedding} refers to the different input embedding vectors that are fed into the CNN-based architecture.}
\centering
\footnotesize
 \begin{tabular}{l l l  l l  l l  l l l}
  \toprule
  \textbf{Input Embedding} & \textbf{Dim} &\multicolumn{2}{c}{Macro} &\multicolumn{2}{c}{Micro} &  \\
  \cmidrule{2-7}
  & & \textbf{F1} & \textbf{AUC} & \textbf{F1} & \textbf{AUC} & \textbf{P@8}\\
  \midrule
  \textbf{BASELINE}\\
  Word2Vec-MIMIC (CAML ~\cite{DBLP:conf/naacl/MullenbachWDSE18})  & 200 & 0.5554	& 0.8767 & 0.6749 & 0.9218 & 0.4034\\
  Word2Vec-MIMIC-MultiResCNN~\cite{li2020icd} & 200 & 0.2519 & 0.7195 & 0.4225 & 0.8126 & 0.3151 \\
  \midrule
  \textbf{POST-PROCESSED EMBEDDINGS} \\
  MeanDiff-Word2Vec-MIMIC & 200 & 0.5525 & 0.8749 & 0.6714 & 0.9205 & 0.4011

 \\
  MeanDiff-PCADiff-Word2Vec-MIMIC & 200 & 0.5325 & 0.8726 & 0.6749 & 0.9214 & 0.4021\\
  \midrule
  \textbf{META-EMBEDDING: Word2Vec-MIMIC,} \\
  \textbf{Word2Vec-PubMed} \\
  Averaging & 200 & 0.5750	& 0.8852
	& 0.6819	& 0.9248 & 0.4074\\
  Locally-linear & 200 & $\textbf{0.5957}$ & $\textbf{0.8940}$ & $\textbf{0.6840}$ & $\textbf{0.9280}$ & $\textbf{0.4088}$\\
  Locally-linear (MultiResCNN) & 200 &  0.4061 & 0.8139 & 0.5678 & 0.8813 & 0.3695 \\
  \midrule
  \textbf{META-EMBEDDINGS on POST-PROCESSED} \\
  \textbf{EMBEDDINGS} \\
  MeanDiff + Averaging & 200 & 0.6144 & 0.9043 & 0.6923 & 0.9336 & 0.4135\\
  MeanDiff-PCADiff + Averaging & 200 & 0.6059 & 0.9004 & 0.6988 & 0.9341 & 0.4132 \\
  MeanDiff + Locally Linear & 200 & \textbf{0.6212} & \textbf{0.9096} & \textbf{0.7043} & \textbf{0.9381} &\textbf{0.4163}\\
  MeanDiff-PCADiff + Locally Linear & 200 & 0.6205 & 0.9080 & 0.7014 & 0.9372 & 0.4160\\

  \bottomrule
 \end{tabular}

\label{resultunstructured32}

\centering
\caption{Micro and Macro results are presented for the multimodal multilabel classification of 32 ICD-10 codes.}
\footnotesize
 \begin{tabular}{l l l  l l  l l  l l }
  \toprule
  \textbf{Unstructured data features} &\multicolumn{2}{c}{Macro} &\multicolumn{2}{c}{Micro} \\
  \cmidrule{2-5}
   & \textbf{F1} & \textbf{AUC} & \textbf{F1} & \textbf{AUC}\\
  \midrule
  \textbf{BASELINE}\\
  - & 0.33211 & 0.6078 & 0.4521	& 0.6608\\
  Word2Vec-MIMIC & 0.5557 &0.7294 & 0.6771 & 0.7921\\	
  \midrule
  \textbf{META-EMBEDDING:} \\
  \textbf{Word2Vec-MIMIC, Word2Vec-PubMed} \\
  Locally linear meta-embedding	& 0.5826 & 0.7972 & 0.6821 & 0.7436 \\
   & $\pm$ 0.0083 & $\pm$ 0.0028 & $\pm$ 0.0023 & $\pm$ 0.0017 \\
  MeanDiff + Locally Linear & \textbf{0.6080} & \textbf{0.7593} & \textbf{0.7057} & \textbf{0.8120} \\
  & $\pm$ 0.0073 & $\pm$ 0.0022 & $\pm$ 0.0022 & $\pm$ 0.0016 \\
  MeanDiff-DiffPCA + Locally Linear & 0.6047 & 0.7577 & 0.7030 & 0.8103 \\
  & $\pm$ 0.0039 & $\pm$ 0.0013 & $\pm$ 0.0025 & $\pm$ 0.0019 \\ 
  \bottomrule
 \end{tabular}
\label{resultstructured32}
\end{table*}
\begin{table*}
\centering
\caption{Micro and macro results are presented for multilabel classification of 50 ICD-9 CODES on the test set. \emph{Input Embedding} refers to the different input embedding vectors that are fed into the CNN-based architecture.}
\footnotesize
 \begin{tabular}{l l l  l l  l l  l l l}
  \toprule
  \textbf{Input Embedding} & \textbf{Dim} &\multicolumn{2}{c}{Macro} &\multicolumn{2}{c}{Micro} & \\
  \cmidrule{2-7}
  & & \textbf{F1} & \textbf{AUC} & \textbf{F1} & \textbf{AUC} & \textbf{P@5}\\
  \midrule
  \textbf{BASELINE}\\
  Word2Vec-MIMIC (CAML ~\cite{DBLP:conf/naacl/MullenbachWDSE18}) & 200 & 0.5571 & 0.8693 & 0.6084 & 0.8910 & 0.5829\\
  Word2Vec-MIMIC-MultiResCNN (\cite{li2020icd}) & 200 & 0.3616 & 0.7643 & 0.5425 & 0.8492 & 0.3415\\
  \midrule
  \textbf{POST-PROCESSED EMBEDDINGS} \\
  MeanDiff-Word2Vec-MIMIC & 200 & 	0.5597 & 0.8620 & 0.5974 & 0.8848 & 0.5739
\\
  MeanDiff-PCADiff-Word2Vec-MIMIC & 200 & 0.5544 & 0.8577 & 0.6022 & 0.8799 & 0.5730 \\
  \midrule
  \textbf{META-EMBEDDING: Word2Vec-MIMIC,} \\
  \textbf{Word2Vec-PubMed} \\
  Averaging & 200 & 0.5663 & 0.8642	& 0.6132 & 0.8894 & 0.5882 \\
  Locally-linear & 200 & \textbf{0.5709} & \textbf{0.8666} & \textbf{0.6161} & \textbf{0.8914} & \textbf{0.5919}\\
  Locally-linear (MultiResCNN) & 200 & 0.4475 & 0.8159 & 0.5761 &  0.8849 & 0.3716 \\
  \midrule
  \textbf{META-EMBEDDINGS on POST-PROCESSED} \\
  \textbf{EMBEDDINGS} \\
  MeanDiff + Averaging & 200 & 0.5665 & 0.8719 & 0.6079 & 0.8937 & 0.5863 \\
  MeanDiff-DiffPCA + Averaging & 200 & 0.5489 & 0.8685 & 0.6049 & 0.8948 & 0.5631\\
  MeanDiff + Locally Linear & 200 & 0.5730 & 0.8758 & \textbf{0.6224} & \textbf{0.9011} & \textbf{0.5732}\\
  MeanDiff-DiffPCA + Locally Linear & 200 & \textbf{0.5740} &  \textbf{0.8760} & 0.6183 & 0.8967 & 0.5661\\

  \bottomrule
 \end{tabular}
\label{resultunstructured50}

\centering
\caption{Micro and Macro results are presented for the multimodal multilabel classification of 50 ICD-9 codes.}
\footnotesize
 \begin{tabular}{l l l  l l  l l  l l }
  \toprule
  \textbf{Unstructured data features} &\multicolumn{2}{c}{Macro} &\multicolumn{2}{c}{Micro} \\
  \cmidrule{2-5}
   & \textbf{F1} & \textbf{AUC} & \textbf{F1} & \textbf{AUC}\\
  \midrule
  \textbf{BASELINE}\\
  - & 0.3940 & 0.6679 & 0.4662 & 0.6417 \\
  Word2Vec-MIMIC & 0.5457 & 0.7179 & 0.6078 & 0.7499\\	
  \midrule
  \textbf{META-EMBEDDING:} \\
  \textbf{Word2Vec-MIMIC, Word2Vec-PubMed} \\
  Locally linear meta-embedding	& 0.5416 & 0.7182 & 0.6077 & 0.7506 \\
  & $\pm$ 0.012 & $\pm$ 0.0046 & $\pm$ 0.0096 & $\pm$ 0.0051 \\
  MeanDiff + Locally Linear & \textbf{0.5495} & \textbf{0.7528} & \textbf{0.6122} & \textbf{0.7528} \\
  & $\pm$ 0.0088 & $\pm$ 0.0059 & $\pm$ 0.012 & $\pm$ 0.0074 \\
  MeanDiff-DiffPCA + Locally Linear & 0.5530 & 0.7215 & 0.6113 & 0.7523 \\
  & $\pm$ 0.010 & $\pm$ 0.0054 & $\pm$ 0.0113 & $\pm$ 0.0076 \\
  \bottomrule
 \end{tabular}
\vspace{-3mm}
\label{resultstructured50}
\end{table*}

Table.~\ref{resultunstructured32} contains the results of experiments aimed at multilabel classification of the top 32 ICD-10 codes using unstructured data (Section.~\ref{sec:proposed_unstructured}). The results indicated that the baseline performance of CAML~\cite{DBLP:conf/naacl/MullenbachWDSE18} was better than the more complex MultiResCNN ~\cite{li2020icd} (when training on less epochs than those reported in the original MultiResCNN paper. Refer to sections \ref{sec:experiment_unstructured} for further discussion). As a result, all remaining experiments utilised the CAML architecture as the base neural model. However, we did observe that the MultiResCNN underwent a significant increase in performance when post-processing was applied to the input embeddings. However, further analysis was required to understand the root cause, which was not in scope for this paper. We propose using an ablation study of the MultiResCNN architecture to understand the causes of the performance increase.

The best performance was achieved by applying the post-processing technique \\ $\textbf{MeanDiff}$ to the PubMed and MIMIC embeddings and then combining them using the the locally linear meta-embedding technique~\cite{DBLP:conf/ijcai/BollegalaHK18}. However, we hypothesised that the post processing step was only effective for the PubMed embeddings and not the MIMIC embeddings. This conclusion was drawn by comparing the baseline experiment (Word2Vec-MIMIC (CAML ~\cite{DBLP:conf/naacl/MullenbachWDSE18})) with the one that only used post-processed MIMIC embeddings (MeanDiff-Word2Vec-MIMIC), and comparing the experiment utilising meta-embeddings without post-processing (Locally-linear) against the one using meta-embeddings and post-processing (MeanDiff + Locally Linear). In the former i.e. Word2Vec-MIMIC (CAML ~\cite{DBLP:conf/naacl/MullenbachWDSE18}) vs MeanDiff-Word2Vec-MIMIC we observed no incremental performance, thus we concluded that post-processing the MIMIC embeddings provided no performance gain. However, when comparing the results of Locally-linear vs MeanDiff + Locally Linear we did observe an incremental performance. This lead us to two possible conclusions:
\begin{itemize}
    \item \textbf{MeanDiff} was only effective for enhancing the performance of meta-embedding techniques
    \item \textbf{MeanDiff} was only effective when applied to the PubMed embeddings
\end{itemize}

 In the case of the latter, we hypothesised this was caused by the inaccurate grammatical and semantic structure of discharge summaries in MIMIC dataset however, validating this was out of scope for the paper. Furthermore, the results indicated that the information captured by the local neighbourhood in both sources was important for boosting the performance. We found that post-processing word embeddings by removing principal components did not provide any improvement. However, investigating the reasons for this was out of scope for this paper.


We found that the meta-embeddings were able to capture the different semantic information from the different sources i.e. the discharge summaries and the external knowledge from PubMed articles. Meta-embedding techniques do not require the raw data used to originally train the embeddings. Instead, they only require the embeddings for training. This provided an efficient way of improving the performance without adding to the complexity of the model.

Similar to Table.~\ref{resultunstructured32}, Table.~\ref{resultunstructured50} contains the results of experiments aimed at multilabel classification of the top 50 ICD-9 codes using unstructured data (Section.~\ref{sec:proposed_unstructured}). We found a similar trend in performance with regard to the different post-processing and meta embeddings combinations. Critically, the highest performing combination for the classification of ICD 10 codes was also the highest performing for the classification of ICD 9 codes. Overall, our proposed approach clearly outperforms the baseline.

Table.~\ref{resultstructured32} contains the results of experiments aimed at multilabel classification of the top 32 ICD-10 codes via a multimodel approach  (as outlined in Section.~\ref{sec:multimodal}). Based on these results, we found that utilising the structured information on its own did not provide good performance in comparison with the baseline. Similarly, the results in Table.~\ref{resultstructured50} indicated that the structured data alone had poor performance but was improved when ensembling with predictions from other models. Crucially, the results indicated that ensembling the structured data with a CNN model using MIMIC and PubMed embeddings that have been processed via $\textbf{MeanDiff}$ and then combined via the locally linear meta-embeddings technique, provided the highest performance of all ensemble models tested.

Our aim was to understand whether we could enhance the prediction of structured information by using our proposed approach for unstructured information. Overall, the results based on the unstructured data indicated that our proposed approach is effective and outperforms the baseline approach. In addition, the results from our proposed approach enhances the performance of the multimodal multi-label classification.

\section{Limitations}
In this work, we did not focus on the interpretability of the results but rather we refer the reader to related work. Specifically, we would like to refer to ~\cite{DBLP:conf/naacl/MullenbachWDSE18} for predictions based on unstructured texts and ~\cite{xu2019multimodal,scheurwegs2016data} for multimodal data. Since, these interpretability methods have been shown to be useful, we make an assumption that their validation results hold for our experiments as well. Furthermore, in this paper we did not attempt to improve the current architecture~\cite{DBLP:conf/naacl/MullenbachWDSE18} but rather focused on investigating the benefit of utilising better features as well as structured data.

We did not explore the use of BERT-based models but refer to prior work~\cite{DBLP:conf/emnlp/ChalkidisFKMAA20}, which indicated why it was not effective to use BERT on the MIMIC dataset. The paper noted a few reasons why BERT models were not effective when applied to the MIMIC dataset namely, the constraint on the word limits i.e. BERT only accepts 512 tokens and, most of the biomedical terms in MIMIC are over fragmented. However, in the future we would like to explore the use of meta-embedding techniques when applied to contextualised word embeddings.

We did not experiment with the full set of ICD codes since these have not been widely used in the current literature. Rather, the top 32 ICD 10 and top 50 ICD 9 codes have been used to report performance. This is a result of the high imbalance of ICD codes present in the MIMIC dataset. Therefore, given the high amount of data required to train ICD classification models, we would not expect models to perform well on ICD codes which rarely appear in the dataset. 

\section{Conclusion and Future Work}
In this paper, we presented a novel multimodal approach for predicting ICD codes using unstructured and structured information. In particular, we studied the effect of infusing external knowledge for enhancing the performance of the current
state-of-the-art model using unstructured information by effectively exploiting the geometric properties of pre-trained word embeddings as well as, combining external knowledge using meta-embedding techniques. We empirically showed that our proposed approach can enhance the performance of current state-of-the-art approaches used for multilabel classification of discharge summaries in the MIMIC-III dataset without relying on more complex architectures or using additional knowledge from the descriptions of ICD codes. In particular, post-processing word vectors and then, combining different pre-trained word embeddings using locally-linear meta-embeddings provided the best performance. In addition, we empirically showed that unstructured information enhanced the performance of the multimodal multi-label classification.

In the future, we would like to investigate softer-measures of performing post-processing of pre-trained word embeddings such as using concept negators~\cite{DBLP:conf/aaai/LiuUS19}. We would also like to investigate the potential to exploit the hierarchical structure of the text present in discharge summaries using hyperbolic-based embedding vectors~\cite{DBLP:conf/aaai/LiuUS19} and also, combine different hyperbolic-based embeddings using meta-embedding techniques~\cite{DBLP:conf/rep4nlp/JawanpuriaDKM20}. Finally, we will experiment with more statistical features and other machine learning algorithms to assess whether we can further improve performance from the structured data.
\bibliographystyle{splncs04}
\bibliography{arxiv_draft}

\begin{thebibliography}{10}
\providecommand{\url}[1]{\texttt{#1}}
\providecommand{\urlprefix}{URL }
\providecommand{\doi}[1]{https://doi.org/#1}

\bibitem{DBLP:conf/ijcai/BollegalaHK18}
Bollegala, D., Hayashi, K., Kawarabayashi, K.: Think globally, embed locally -
  locally linear meta-embedding of words. In: Proceedings of IJCAI. pp.
  3970--3976 (2018)

\bibitem{DBLP:conf/emnlp/ChalkidisFKMAA20}
Chalkidis, I., Fergadiotis, M., Kotitsas, S., Malakasiotis, P., Aletras, N.,
  Androutsopoulos, I.: An empirical study on large-scale multi-label text
  classification including few and zero-shot labels. In: Webber, B., Cohn, T.,
  He, Y., Liu, Y. (eds.) Proceedings of the 2020 Conference on Empirical
  Methods in Natural Language Processing, {EMNLP} 2020, Online, November 16-20,
  2020. pp. 7503--7515. Association for Computational Linguistics (2020).
  \doi{10.18653/v1/2020.emnlp-main.607},
  \url{https://doi.org/10.18653/v1/2020.emnlp-main.607}

\bibitem{DBLP:conf/biostec/ChowdhuryZYL20}
Chowdhury, S., Zhang, C., Yu, P.S., Luo, Y.: Med2meta: Learning representations
  of medical concepts with meta-embeddings. In: Proceedings of HEALTHINF. pp.
  369--376 (2020)

\bibitem{DBLP:conf/naacl/CoatesB18}
Coates, J., Bollegala, D.: Frustratingly easy meta-embedding - computing
  meta-embeddings by averaging source word embeddings. In: Proceedings of
  NAACL-HLT. pp. 194--198 (2018)

\bibitem{el2019embedding}
El~Boukkouri, H., Ferret, O., Lavergne, T., Zweigenbaum, P.: Embedding
  strategies for specialized domains: Application to clinical entity
  recognition. In: Proceedings of ACL. pp. 295--301 (2019)

\bibitem{DBLP:conf/rep4nlp/JawanpuriaDKM20}
Jawanpuria, P., Dev, N.T.V.S., Kunchukuttan, A., Mishra, B.: Learning geometric
  word meta-embeddings. In: Proceedings of RepL4NLP@ACL. pp. 39--44 (2020)

\bibitem{mimiciii}
Johnson, A.E., Pollard, T.J., Shen, L., Li-wei, H.L., Feng, M., Ghassemi, M.,
  Moody, B., Szolovits, P., Celi, L.A., Mark, R.G.: Mimic-iii, a freely
  accessible critical care database. Scientific data  \textbf{3},  160035
  (2016)

\bibitem{DBLP:conf/emnlp/KielaWC18}
Kiela, D., Wang, C., Cho, K.: Dynamic meta-embeddings for improved sentence
  representations. In: Proceedings of EMNLP. pp. 1466--1477 (2018)

\bibitem{DBLP:conf/emnlp/Kim14}
Kim, Y.: Convolutional neural networks for sentence classification. In:
  Proceedings of EMNLP. pp. 1746--1751 (2014)

\bibitem{koopman2015automatic}
Koopman, B., Zuccon, G., Nguyen, A., Bergheim, A., Grayson, N.: Automatic
  icd-10 classification of cancers from free-text death certificates.
  International journal of medical informatics  \textbf{84}(11),  956--965
  (2015)

\bibitem{li2020icd}
Li, F., Yu, H.: Icd coding from clinical text using multi-filter residual
  convolutional neural network. In: Proceedings of AAAI. pp. 8180--8187 (2020)

\bibitem{DBLP:conf/iclr/LinFSYXZB17}
Lin, Z., Feng, M., dos Santos, C.N., Yu, M., Xiang, B., Zhou, B., Bengio, Y.: A
  structured self-attentive sentence embedding. In: Proceedings of ICLR (2017)

\bibitem{DBLP:conf/aaai/LiuUS19}
Liu, T., Ungar, L., Sedoc, J.: Unsupervised post-processing of word vectors via
  conceptor negation. In: Proceedings of AAAI. pp. 6778--6785 (2019)

\bibitem{mikolov2013distributed}
Mikolov, T., Sutskever, I., Chen, K., Corrado, G.S., Dean, J.: Distributed
  representations of words and phrases and their compositionality. In:
  Proceesings of NIPS. pp. 3111--3119 (2013)

\bibitem{moen2013distributional}
Moen, S., Ananiadou, T.S.S.: Distributional semantics resources for biomedical
  text processing. Proceedings of LBM pp. 39--44 (2013)

\bibitem{DBLP:conf/iclr/MuV18}
Mu, J., Viswanath, P.: All-but-the-top: Simple and effective postprocessing for
  word representations. In: Proceedings of ICLR (2018)

\bibitem{DBLP:conf/naacl/MullenbachWDSE18}
Mullenbach, J., Wiegreffe, S., Duke, J., Sun, J., Eisenstein, J.: Explainable
  prediction of medical codes from clinical text. In: Proceedings of NAACL-HLT.
  pp. 1101--1111 (2018)

\bibitem{pennington2014glove}
Pennington, J., Socher, R., Manning, C.D.: Glove: Global vectors for word
  representation. In: Proceedings of EMNLP. pp. 1532--1543 (2014)

\bibitem{scheurwegs2016data}
Scheurwegs, E., Luyckx, K., Luyten, L., Daelemans, W., Van~den Bulcke, T.: Data
  integration of structured and unstructured sources for assigning clinical
  codes to patient stays. Journal of the American Medical Informatics
  Association  \textbf{23}(e1),  e11--e19 (2016)

\bibitem{Shi2017TowardsAI}
Shi, H., Xie, P., Hu, Z., Zhang, M., Xing, E.: Towards automated icd coding
  using deep learning. ArXiv  \textbf{abs/1711.04075} (2017)

\bibitem{ijcai2020-461}
Vu, T., Nguyen, D.Q., Nguyen, A.: A label attention model for icd coding from
  clinical text. In: Proceedings of IJCAI. pp. 3335--3341 (2020)

\bibitem{xu2019multimodal}
Xu, K., Lam, M., Pang, J., Gao, X., Band, C., Mathur, P., Papay, F., Khanna,
  A.K., Cywinski, J.B., Maheshwari, K., et~al.: Multimodal machine learning for
  automated icd coding. In: Proceedings of Machine Learning for Healthcare
  Conference. pp. 197--215 (2019)

\bibitem{yin2016learning}
Yin, W., Sch{\"u}tze, H.: Learning word meta-embeddings. In: Proceedings of
  ACL. pp. 1351--1360 (2016)

\end{thebibliography}

\end{document}